\documentclass{llncs}
\usepackage[utf8]{inputenc}

\usepackage{epsfig}
\usepackage{placeins}
\usepackage{multirow}
\usepackage{array}
\usepackage{subfigure}
\usepackage{bbding}

\title{Multimodal Registration of Retinal Images Using Domain-Specific Landmarks and Vessel Enhancement}

\author{Álvaro S. Hervella\inst{1,2}$^\textrm{\Envelope}$ \and José Rouco\inst{1,2} \and Jorge Novo\inst{1,2} \and Marcos Ortega\inst{1,2}}

\institute{CITIC-Research Center of Information and Communication Technologies, University of A Coruña, A Coruña (Spain) \and Department of Computer Science, University of A Coruña, A Coruña (Spain) \\ \email{a.suarezh@udc.es}}

\date{November 2017}

\begin{document}

\maketitle

\begin{abstract}
The analysis of different image modalities is frequently performed in ophthalmology as it provides complementary information for the diagnosis and follow-up of relevant diseases, like hypertension or diabetes. This work presents a hybrid method for the multimodal registration of color fundus retinography and fluorescein angiography. The proposed method combines a feature-based approach, using domain-specific landmarks, with an intensity-based approach that employs a domain-adapted similarity metric. The methodology is tested on a dataset of 59 image pairs containing both healthy and pathological cases. The results show a satisfactory performance of the proposed combined approach in this multimodal scenario, improving the registration accuracy achieved by the feature-based and the intensity-based approaches.
\end{abstract}

\section{Introduction}
Multimodal medical image registration is important in the context of diagnosis and follow-up of many relevant diseases. An accurate multimodal registration allows the integration of information obtained from different image modalities, providing complementary information to the clinicians, and improving the diagnostic capabilities. Ophthalmology benefits from this fact given the significant number of existing retinal image modalities: color fundus retinography, fluorescein angiography, autofluorescence fundus retinography or red-free fundus retinography, among others. These modalities offer different visualizations of the eye fundus anatomical structures, lesions, and pathologies, without the possibility of achieving the combined multimodal information using only one of the modalities.

In general, registration algorithms can be classified in two groups: feature-based registration (FBR) and intensity-based registration (IBR) \cite{medical_review}. 
FBR methods use interest points, such as landmarks, along with local shape features of their neighborhoods to find point correspondences and estimate the spatial transformation between the images. For the detection of interest points, common algorithms as Harris corner detector \cite{piifd}, SIFT \cite{sift} \cite{quadratic}, SURF \cite{surf} as well as variations of them \cite{ur-sift} \cite{edgemap}, have been used in different proposals for retinal images. These algorithms detect a large number of interest points in the images. However, as the detected points are not necessarily representative characteristics of the retinal image contents, many of them may not be present across the different modalities. An excessive number of non-representative detected points increases the computational cost of the posterior point matching and increases the likelihood of matching wrong correspondences. On the other hand, the application of generic local shape descriptors is also limited by the differences among retinal image modalities, and usually requires preprocessing for multimodal scenarios \cite{edgemap}. Some proposals solve this issue with the design of domain-specific descriptors \cite{lospa} \cite{piifd}, but they still rely on non-specific methods for the detection of interest points. The use of algorithms that aim at detecting common retinal structures can provide more representative and repeatable characteristics. The detection of line structures, which usually correspond to vessels and disease boundaries, may be seen as a first approach to detect representative characteristics of the retinal images \cite{lines}. However, these characteristics may not have a clear correspondence in all modalities. More representative characteristics can be obtained with the extraction of natural landmarks, such as vessel intersections. This idea was tested by Lalibert\'e et al. \cite{central} for the registration of retinal images although, their method, that also requires the detection of the optic disk, was not robust enough and failed in several images. The use of these natural landmarks is not explored in posterior works, to our knowledge, even though its successful application can greatly reduce the number of detected candidate points for the matching process. 

IBR methods use similarity metrics that take into account the intensity values of the whole images instead of sparse local neighborhoods. This allows performing the registration with high order transformations \cite{medical_review}, as it prevents the risk of overfitting to a small number of points. The registration is performed by optimizing a similarity measure, as intensity differences or cross-correlation for monomodal cases, or mutual information (MI) for multimodal cases. Nevertheless, the application in multimodal scenarios depends on the complexity of the image modalities and the relation between their intensity distributions. Specifically for retinal images, Legg et al. \cite{improving_mi} found that in some cases there is an inconsistency between the MI value and the accuracy of the registration, existing transformations with better MI scores than the ground truth registration. These difficulties may explain the reduced number of IBR proposals for multimodal retinal image registration.
Another use of the IBR approach is in combination with FBR methods, being combined in hybrid methodologies that try to exploit the capabilities of both strategies \cite{hybrid_retinal} \cite{hybrid_pc}. 

In this work, we propose a hybrid methodology for the multimodal registration of color fundus retinographies and fluorescein angiographies. The method combines an initial FBR approach driven by domain-specific landmarks, with an IBR refinement that uses a domain-adapted similarity metric. Both approaches exploit the presence of the retinal vascular tree in the retinographies and the angiographies. The proposed FBR is based on the detection of landmarks present in both retinal image modalities, i.e. vessel bifurcations and crossovers. These landmarks can be detected with high specificity, which greatly reduces the number of detected points and facilitates the subsequent point matching. We completely avoid the descriptor computational step, as the point matching with the geometric information only is enough to find the correspondences between the two images. The latter IBR aims to refine the registration through the estimation of a high order transformation. To perform the IBR over the multimodal images, a domain adapted similarity metric is used. This adaptation consists in the enhancement of vessel regions, and transforms retinography and angiography to a common image space where the similarity metrics from the monomodal scenarios can be employed. Experiments are conducted to evaluate the performance of the hybrid approach and the achieved improvement with respect to the independent application of the FBR and IBR methods.

\section{Methodology} \label{methodology}

\subsection{Landmark-based Registration}
\label{fbr}
The retinal vascular tree is a complex network of arteries and veins that frequently branch and intersect. The intersection points of the blood vessel segments are natural characteristic points of the retina and have proven to be a reliable biometric pattern \cite{minucias}. These intersection points, consisting of vessel bifurcations and crossovers, are used as landmarks. The detection and matching of these domain-specific landmarks is performed following an approach originally proposed for retinal biometric authentication \cite{minucias}. The original method was applied in a monomodal scenario with optic disc centered images to compute the similarity between the vascular trees in two retinographies. The multimodal registration poses the reverse problem, as it is known that both images belong to the same individual and the similarity between them must be maximized. This implies that a higher accuracy in the localization of the landmarks is needed. The mentioned method is adapted to detect landmarks in both retinography and angiography with specific modality modifications.

\begin{figure}[tb]
    \centering
    \subfigure[]{\includegraphics[width=0.4\textwidth]{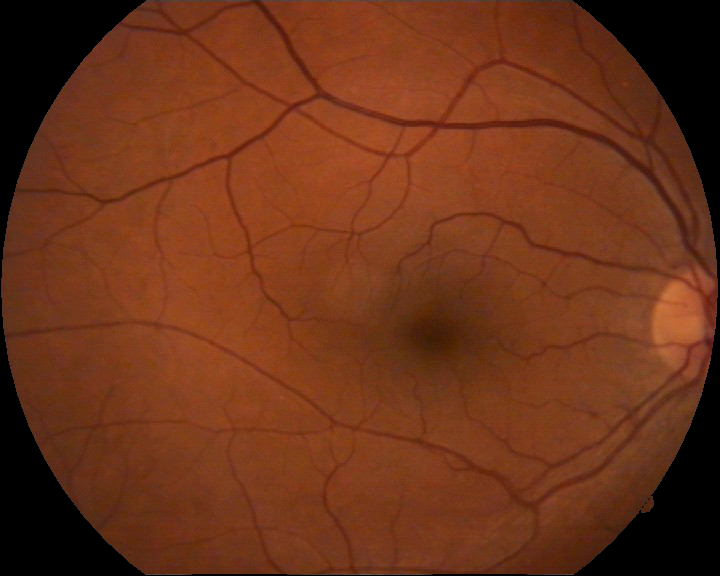}}
    \ \ \
    \subfigure[]{\includegraphics[width=0.4\textwidth]{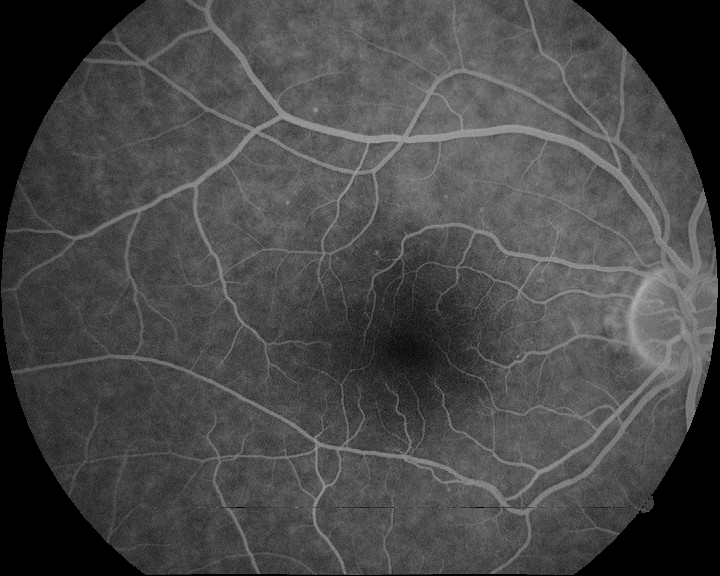}}
    \subfigure[]{\includegraphics[width=0.4\textwidth]{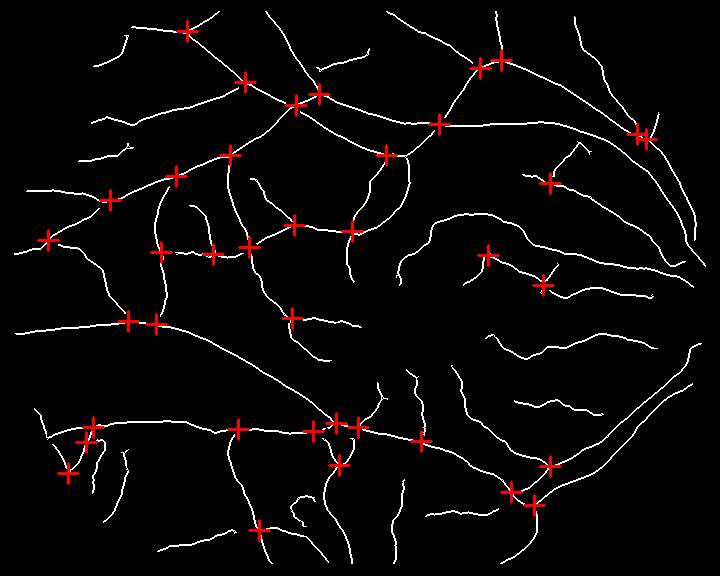}}
    \ \ \
    \subfigure[]{\includegraphics[width=0.4\textwidth]{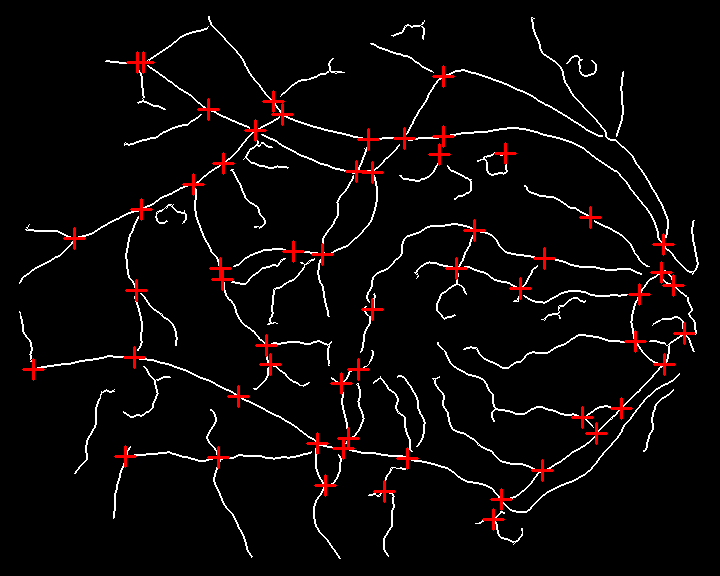}}
    \caption{Example of multimodal image pair and the results of the landmark detection method: (a) input retinography; (b) input angiography; (c)(d) resulting binary vessel trees along with the detected landmarks for (c) the retinography and (d) the angiography.}
    \label{examples}
\end{figure}

Retinal images can be seen as landscapes where vessels appear as creases (ridges and valleys). In retinographies, the vessels are valleys in the landscape while in angiographies they represent ridges. Defining the images as level functions, valleys (or ridges) are formed in the points of negative minima (or positive maxima) curvature. 
The local curvature minima and maxima are detected using the MLSEC-ST operator \cite{mlsec}. The vessel tree is given by the set of valleys (or ridges) for retinography (or angiography). The result is a binary image for each modality, consisting of 1 pixel width vessel segments,

The obtained vessel tree is fragmented at some points. Discontinuities appear at crossovers and bifurcations where vessels with different directions meet, and in the middle of a single vessel due to illumination and contrast variations of the image. Bifurcation and crossover detection is approached by joining the segmented vessels as described in \cite{minucias2}. 
Bifurcations are established where an extended segment under a given maximum distance intersects another segment. Crossovers, instead, are considered as nearby double bifurcations. They are detected at positions where two bifurcations are closer than a given distance and the relative angle between their directions is below a certain threshold. Figure. \ref{examples} shows an example of a retinography/angiography pair and the result of the identified vessel tree and landmarks.

This detection method results in a low number of suitable detected points, allowing to immediately perform the transformation estimation without an additional computation of descriptors.
Bifurcations and crossovers are used to estimate the transformation between image pairs with a point matching algorithm \cite{minucias}. The applied transformation is a restricted form of affine transformation that only considers translation, rotation, and isotropic scaling. Therefore, the transformation has 4 degrees of freedom and can be computed with only two pairs of matched points. For each previously detected bifurcation or crossover, the position and vessel orientation are known as they are directly obtained from the detection method. These two characteristics are enough to perform the registration, without the need of a sophisticated descriptor computation stage. The high specificity of the detection method leads to a low number of detected landmarks per image. Thus, it is practical to consider all the possible matching pairs. The number of possibilities is additionally reduced by taking into account a maximum and minimum scaling factor, which can be computed in advance as the ratio between the distance of two points in an image and the distance of any other two points in the other image. The relative angles between points, derived from the vessel orientation, are also used for additional restrictions \cite{minucias}.

\subsection{Intensity-based Registration} \label{ibr}
The registration accuracy of the proposed FBR method is limited by the low complexity of the transformation considered and the landmark localization precision. A refinement stage that considers high order transformations and pixel-level details is proposed to improve the registration accuracy. In order to estimate higher order transformations, it is convenient to use an IBR approach considering all the pixels of the image pairs. However, it is necessary to design and IBR approach that is well-suited for the target multimodal scenario. To that end, a new domain-adapted similarity metric is constructed combining a vessel enhancement preprocessing with the normalized cross-correlation (NCC) similarity metric. The vessel enhancement transforms images from both modalities to a common image space where the NCC can be successfully employed. This whole operation is named as VE-NCC.

\begin{figure}[tb]
    \centering
    \subfigure[]{\includegraphics[width=0.4\textwidth]{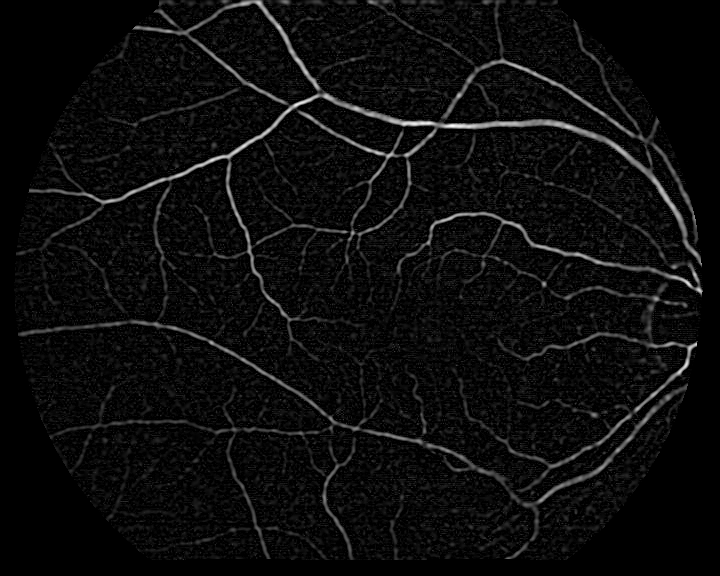}}
    \ \ \
    \subfigure[]{\includegraphics[width=0.4\textwidth]{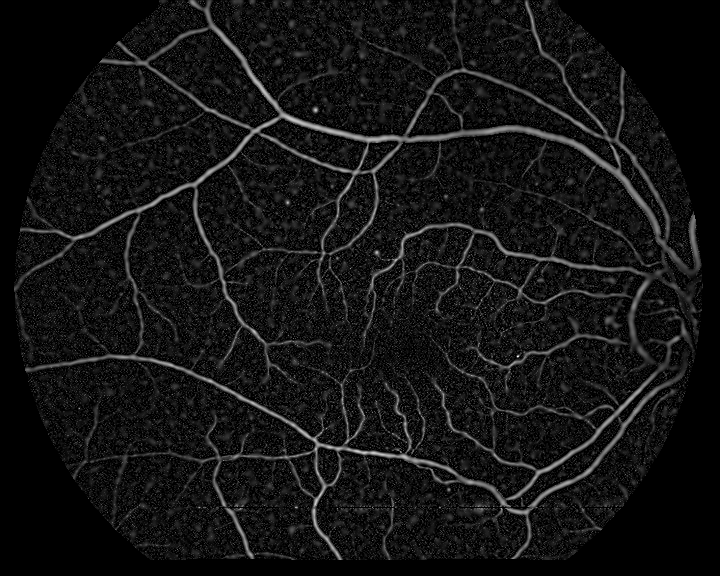}}
    \caption{Examples of the vessel enhancement operation applied to the multimodal image pair in Fig. \ref{examples}: (a) result from retinography; (b) result from angiography.}
    \label{mslp}
\end{figure}

The vessel enhancement is motivated by the fact that the vessels are present in both the retinography and the angiography as tubular regions of low or high intensity values, respectively. These vessels vary in thickness throughout the image and can appear in any direction. This motivates the use of a multiscale analysis that is sensitive to multiple orientations. A scale-space is defined as $I(x,y;t) = I(x,y)*G(x,y;t)$ where $t$ is the scale parameter and $G$ is a Gaussian kernel \cite{lindeberg}. The enhancement of the vessel regions is performed using the Laplace operator $\nabla^2$. The Laplacian image, $\nabla^2I$, will have a high response at nearby positions of the image edges, like those at the vessel boundaries. The distance from the Laplacian peaks to the edges depends on the scale used to compute $\nabla^2I$. The vessel centerlines achieve the maximum response at the scales where the peaks from both vessel boundaries concur. Thus, the scale parameter $t$ allows to control the scale of the vessels to enhance and the maximum Laplacian response along the scale allows to enhance the vessel centerlines. The normalized Laplacian scale-space is defined as:
\begin{equation}
L(x,y;t)=t^2 \nabla^2I(x,y;t) 
\end{equation}
where $t^2$ represents the normalization factor. A property of scale-space representations is that the amplitude of spatial derivatives decreases with the scale \cite{lindeberg}. The normalization factor allows the comparison and combination of the magnitude at different scales. Finally, the maximum value across scales for every point is computed as:
\begin{equation}
L(x,y)=max_{t \in S} \lceil m L(x,y;t) \rceil_{\emptyset}
\end{equation}
where $m=1$ for the angiography and $m=-1$ for the retinography, and $\lceil \cdot \rceil_\emptyset$ denotes the halfwave rectification. The rectification is used to avoid the negative Laplacian peaks outside the vessel regions so that only the vessel interiors are represented in the enhanced images. This results in a common representation for the retinography and the angiography, with enhanced vessel regions and the same intensity level pattern. Figure. \ref{mslp} shows the result of the vessel enhancement operation applied to the retinography/angiography pair exposed in Fig. \ref{examples}.

The transformation between images is obtained through the minimization of the negative VE-NCC. It is important to initialize the algorithm with a proper initial transformation. The estimated transformation from the FBR serves as initialization for the IBR. Two different transformation models are considered to perform the IBR: Affine Transformation (AT) and Free Form Deformation (FFD). AT allows translation, rotation, anisotropic scaling, and shearing, having 6 degrees of freedom. Differently, FFD uses a grid of control points that are moved individually along the image to define a high order transformation.

\section{Experiments and Results} \label{results}

For the evaluation of the proposed methodology, we use the publicly available Isfahan MISP dataset, consisting of retinography and angiography images of diabetic patients \cite{imisp}. This dataset consists of 59 image pairs divided into two collections from healthy and pathological cases. The pathological cases correspond to patients with mild and moderate retinal diseases caused by diabetic retinopathy. The images have a resolution of 720 $\times$ 576 pixels. The division of the dataset in healthy and pathological cases allows analyzing the effect of the pathologies in the registration performance.

\begin{figure}[tb]
    \centering
    \subfigure[]{\includegraphics[width=0.48\textwidth]{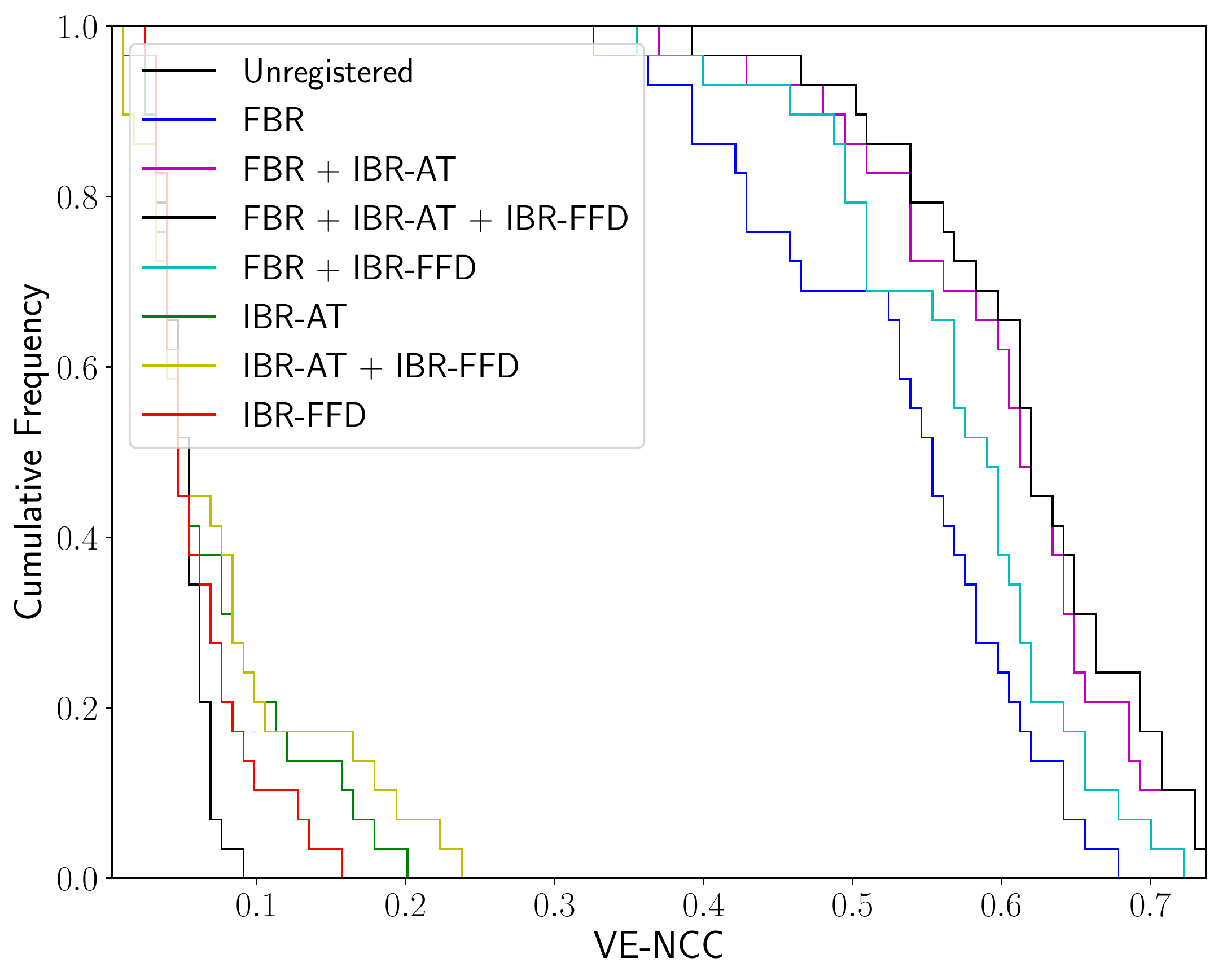}}
    \subfigure[]{\includegraphics[width=0.48\textwidth]{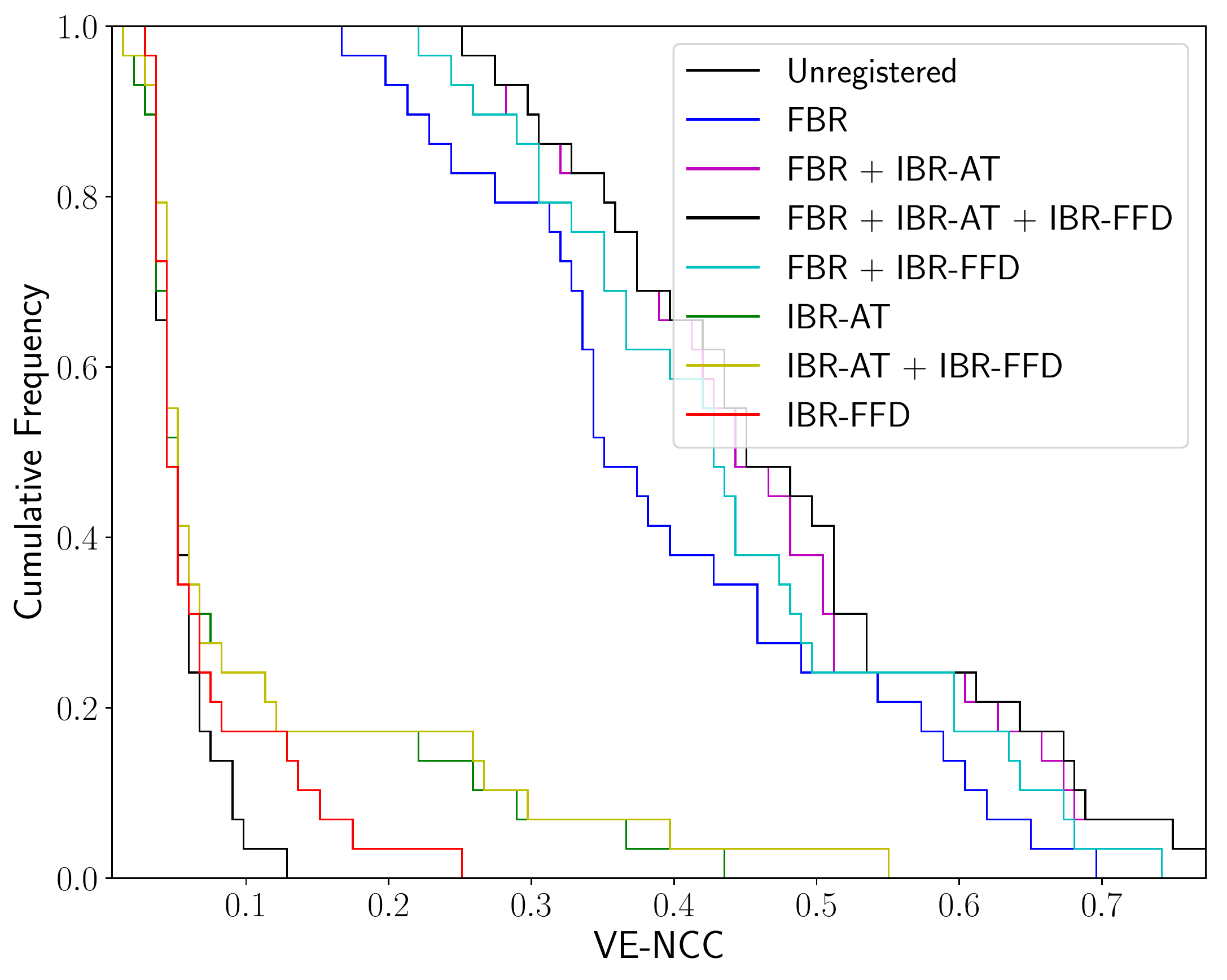}}
    \caption{Cumulative distribution of the VE-NCC: (a) healthy cases: (b) pathological cases.}
    \label{hist}
\end{figure}

\begin{figure}[tb]
    \centering
    \subfigure[]{\includegraphics[width=0.24\textwidth]{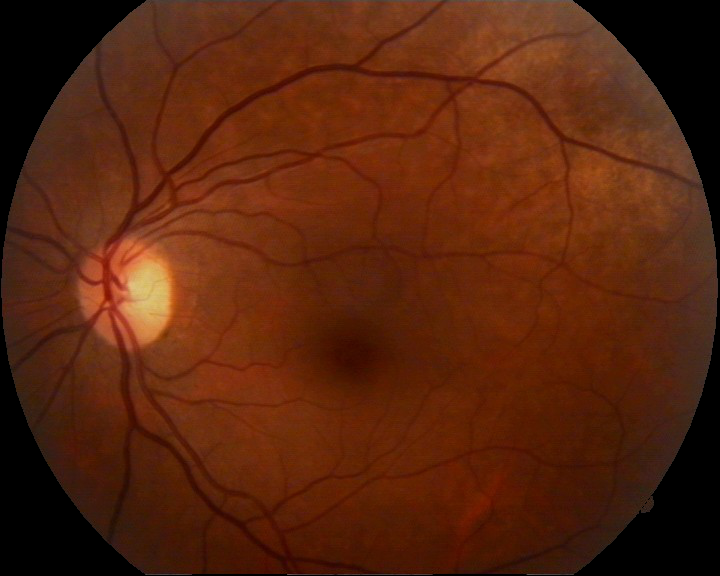}}
    \subfigure[]{\includegraphics[width=0.24\textwidth]{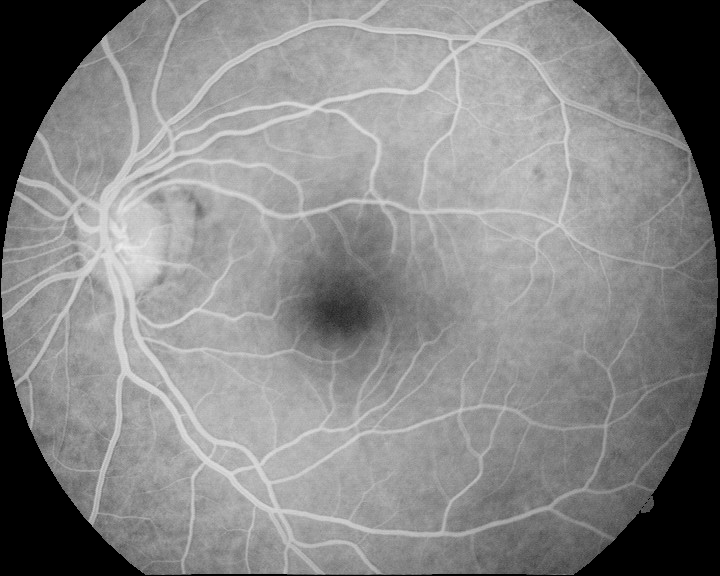}}
    \subfigure[]{\includegraphics[width=0.24\textwidth]{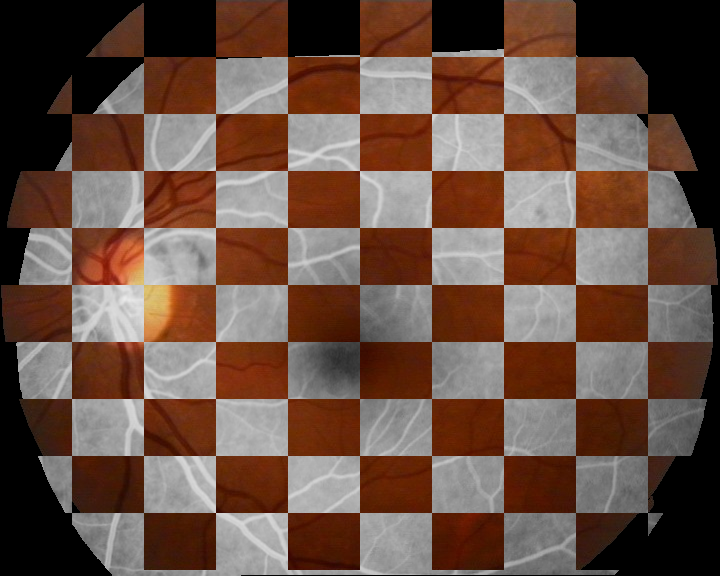}}
    \subfigure[]{\includegraphics[width=0.24\textwidth]{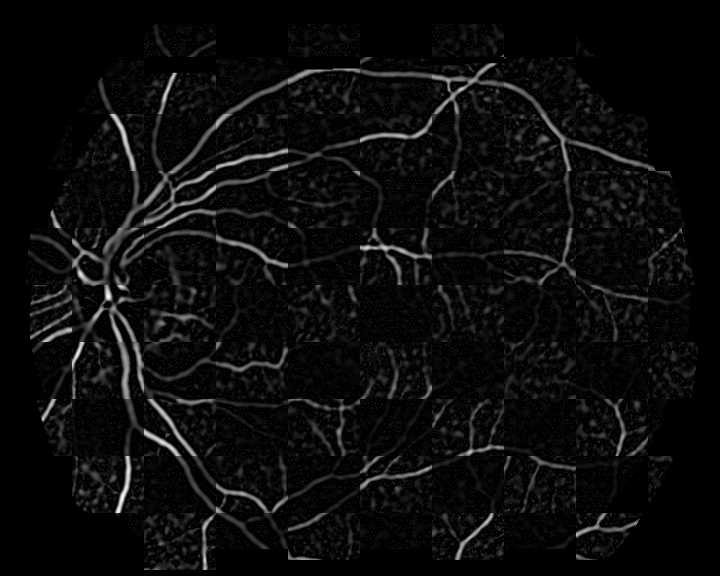}}
    \subfigure[]{\includegraphics[width=0.24\textwidth]{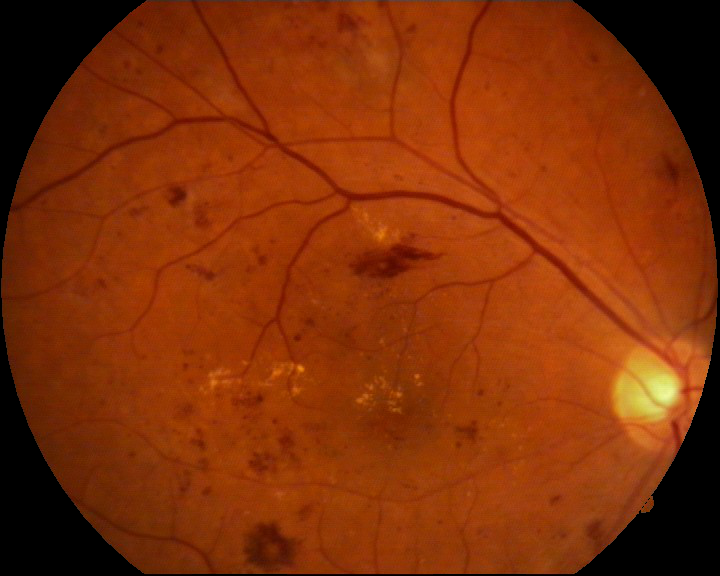}}
    \subfigure[]{\includegraphics[width=0.24\textwidth]{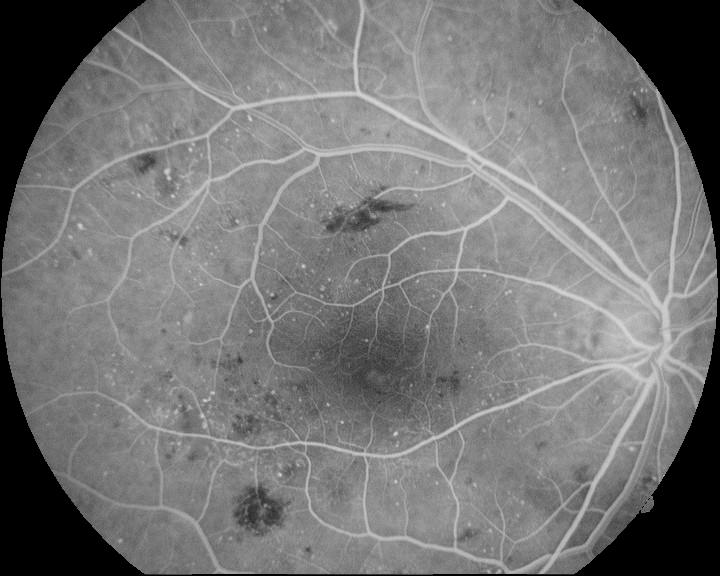}}
    \subfigure[]{\includegraphics[width=0.24\textwidth]{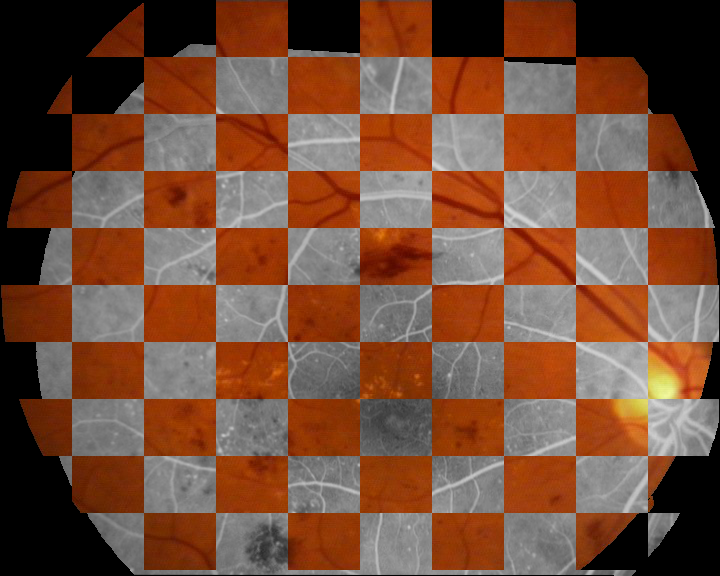}}
    \subfigure[]{\includegraphics[width=0.24\textwidth]{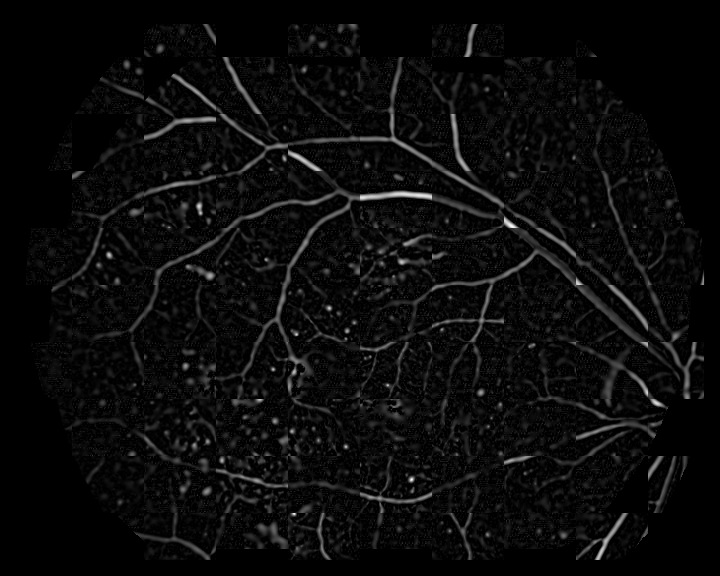}}
    \subfigure[]{\includegraphics[width=0.24\textwidth]{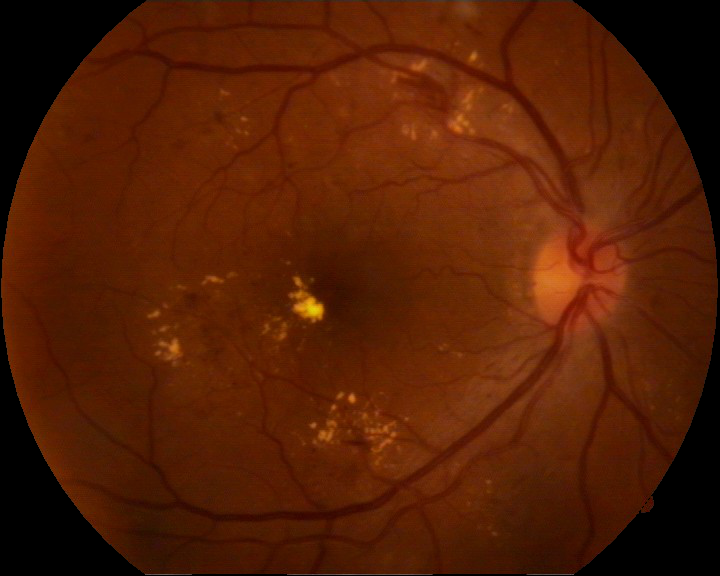}}
    \subfigure[]{\includegraphics[width=0.24\textwidth]{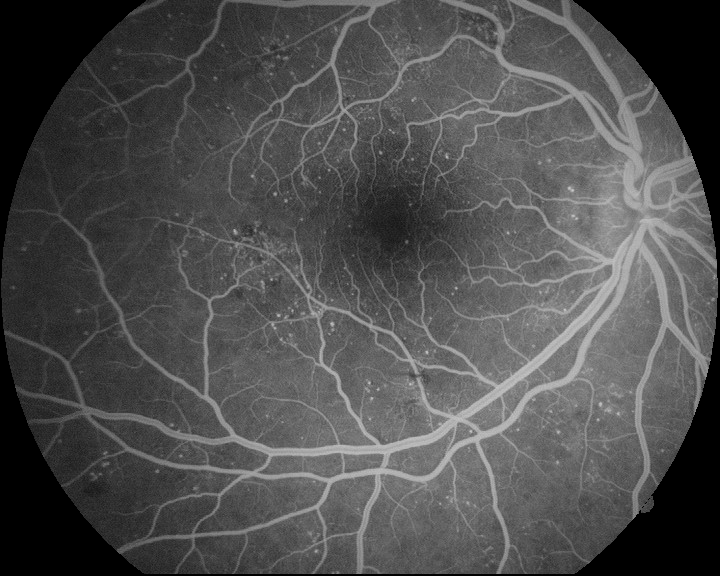}}
    \subfigure[]{\includegraphics[width=0.24\textwidth]{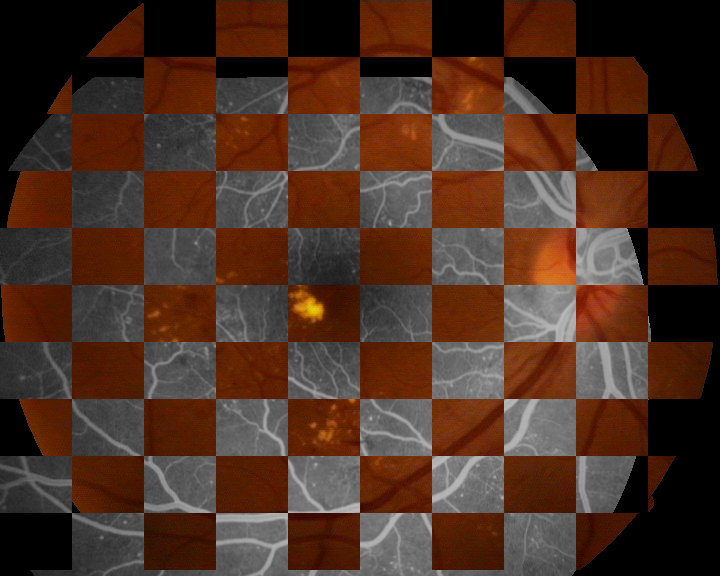}}
    \subfigure[]{\includegraphics[width=0.24\textwidth]{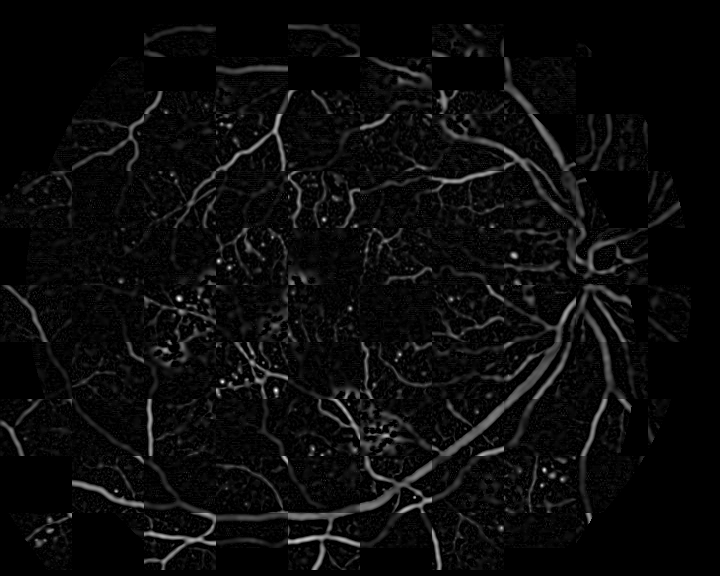}}
    \caption{Examples of the multimodal registration with the hybrid method: (a)(c)(i) retinographies; (b)(f)(j) angiographies; (c)(g)(k) registered image pairs; (d)(h)(l) results of the vessel enhancement operation applied to the registered image pairs.}
    \label{registered_examples}
\end{figure}

Several experiments are conducted to evaluate the hybrid methodology as well as the performance of the FBR and IBR methods. Regarding the IBR method, both the affine transformation (IBR-AT) and the free form deformation (IBR-FFD) variations are applied. We propose the hybrid method formed by the sequential application of FBR, IBR-AT and IBR-FFD, and alternative variations over this by removing one or two steps at a time. This results in 7 different configurations of methods, as reported in Table \ref{experiments}. 
The table shows the average and standard deviation VE-NCC for each method in healthy and pathological cases. Figure \ref{hist} depicts the cumulative distribution of the VE-NCC values. The best results are achieved by the proposed hybrid method. There is a large difference between the experiments that perform the initial FBR and the ones that directly apply IBR. For the latter experiments the registration failed in most of the cases. Most of the image pairs do not significantly change their VE-NCC values by applying IBR alone, and only a few of them obtained values over the minimum that was achieved by the initial FBR. These results indicate that, with the use of IBR and high order transformations, more accurate registrations can be achieved. However, they also evidence the importance of a proper initialization for the convergence of the optimization algorithm, which is provided by the initial FBR. Moreover, the IBR-FFD also benefits from the previous IBF-AT, as the order of the applied transformation directly fixes the search space dimensionality, increasing the complexity of the optimization.
Figure \ref{registered_examples} exposes some representative examples of the images registered with the proposed hybrid method. Both the raw images and the vessel enhanced images provide qualitative evidence of a satisfactory multimodal registration with the hybrid approach in healthy and pathological scenarios.

\begin{table}[tb]
\centering
\caption{Comparison results of average VE-NCC and standard deviation for the different configurations tested.}
\label{experiments}
\begin{tabular*}{\hsize}{@{\extracolsep{\fill}} ccccc }
\hline
\multicolumn{1}{c}{\multirow{2}{*}{FBR}} & \multirow{2}{*}{IBR-AT} & \multirow{2}{*}{IBR-FFD} & \multicolumn{2}{c}{VE-NCC}                                                    \\ \cline{4-5} 
\multicolumn{1}{c}{}                     &                         &                          & \multicolumn{1}{c}{Healthy cases}    & \multicolumn{1}{c}{Phatological cases} \\ \hline

\multicolumn{1}{c}{\textbullet}          & \multicolumn{1}{c}{\textbullet}& \multicolumn{1}{c}{\textbullet}& \multicolumn{1}{c}{\textbf{0.6123$~\pm~$ 0.0815}}                     & \multicolumn{1}{c}{\textbf{0.4758$~\pm~$ 0.1419}}                       \\
\multicolumn{1}{c}{\textbullet}          & \multicolumn{1}{c}{\textbullet} &                          & \multicolumn{1}{c}{0.5980$~\pm~$ 0.0865}  & \multicolumn{1}{c}{0.4661$~\pm~$ 0.1406}   \\
\multicolumn{1}{c}{\textbullet}          &                         & \multicolumn{1}{c}{\textbullet}& \multicolumn{1}{c}{0.5668$~\pm~$ 0.0828}                     & \multicolumn{1}{c}{0.4401$~\pm~$ 0.1381}                       \\

\multicolumn{1}{c}{\textbullet}          &                         &                          & \multicolumn{1}{c}{0.5266$~\pm~$ 0.0928} & \multicolumn{1}{c}{0.3961$~\pm~$ 0.1416}   \\
                                         & \multicolumn{1}{c}{\textbullet}&                          & \multicolumn{1}{c}{0.0673$~\pm~$ 0.0500}                     & \multicolumn{1}{c}{0.0930$~\pm~$ 0.1065}                       \\
                                         & \multicolumn{1}{c}{\textbullet}& \multicolumn{1}{c}{\textbullet}& \multicolumn{1}{c}{0.0733$~\pm~$ 0.0627}                     & \multicolumn{1}{c}{0.1005$~\pm~$ 0.1250}                       \\
                                         &                         & \multicolumn{1}{c}{\textbullet}& \multicolumn{1}{c}{0.0581$~\pm~$ 0.0323}                     & \multicolumn{1}{c}{0.0656$~\pm~$ 0.0497}                       \\ 
                                         \multicolumn{1}{c}{}                     &                         &                          & \multicolumn{1}{c}{0.0481$~\pm~$ 0.0159} & \multicolumn{1}{c}{0.0518$~\pm~$ 0.0220}   \\\hline
\end{tabular*}
\end{table}

\begin{figure}[tb]
    \centering
    \subfigure[]{\includegraphics[width=0.32\textwidth]{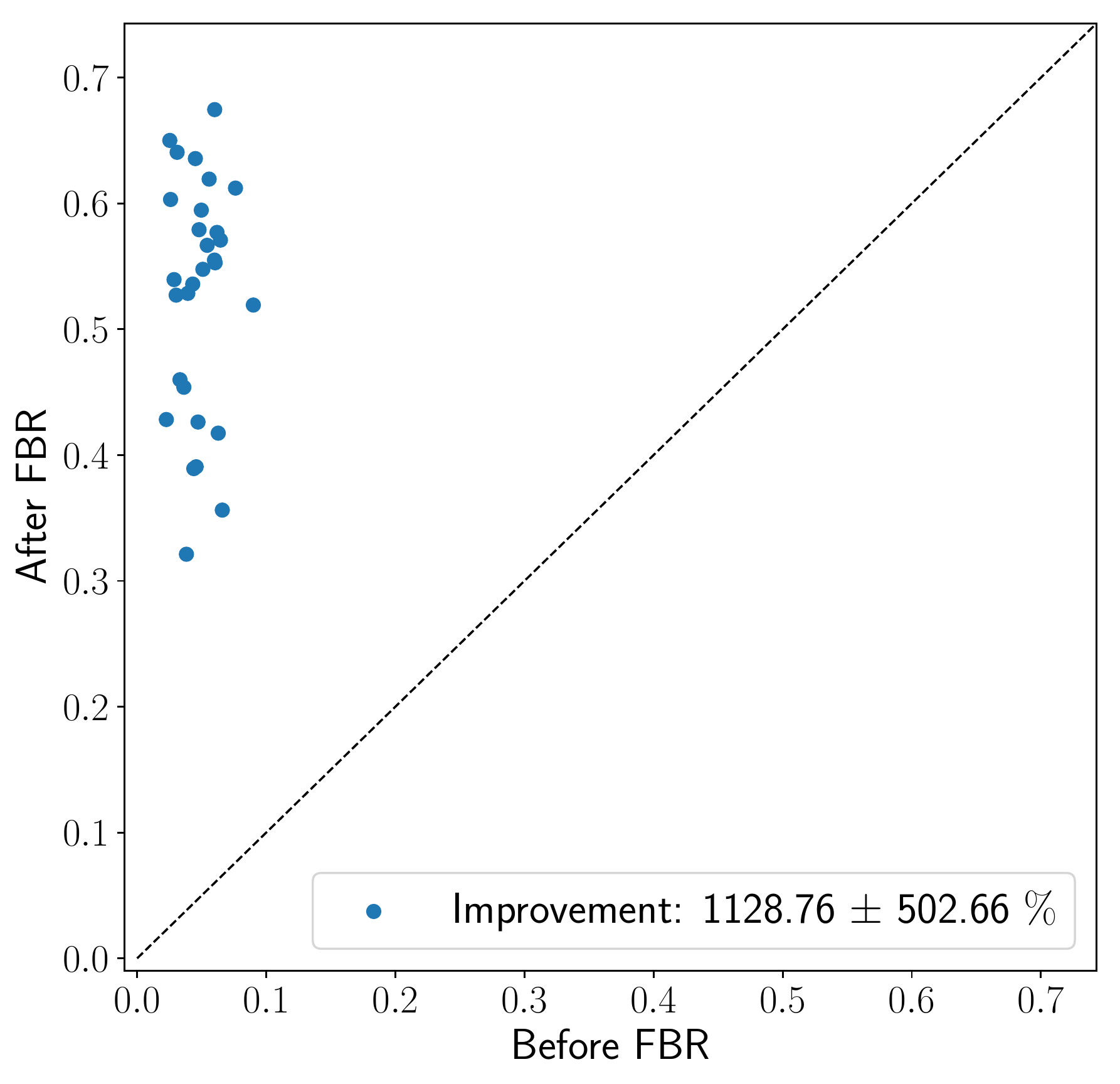}}
    \subfigure[]{\includegraphics[width=0.32\textwidth]{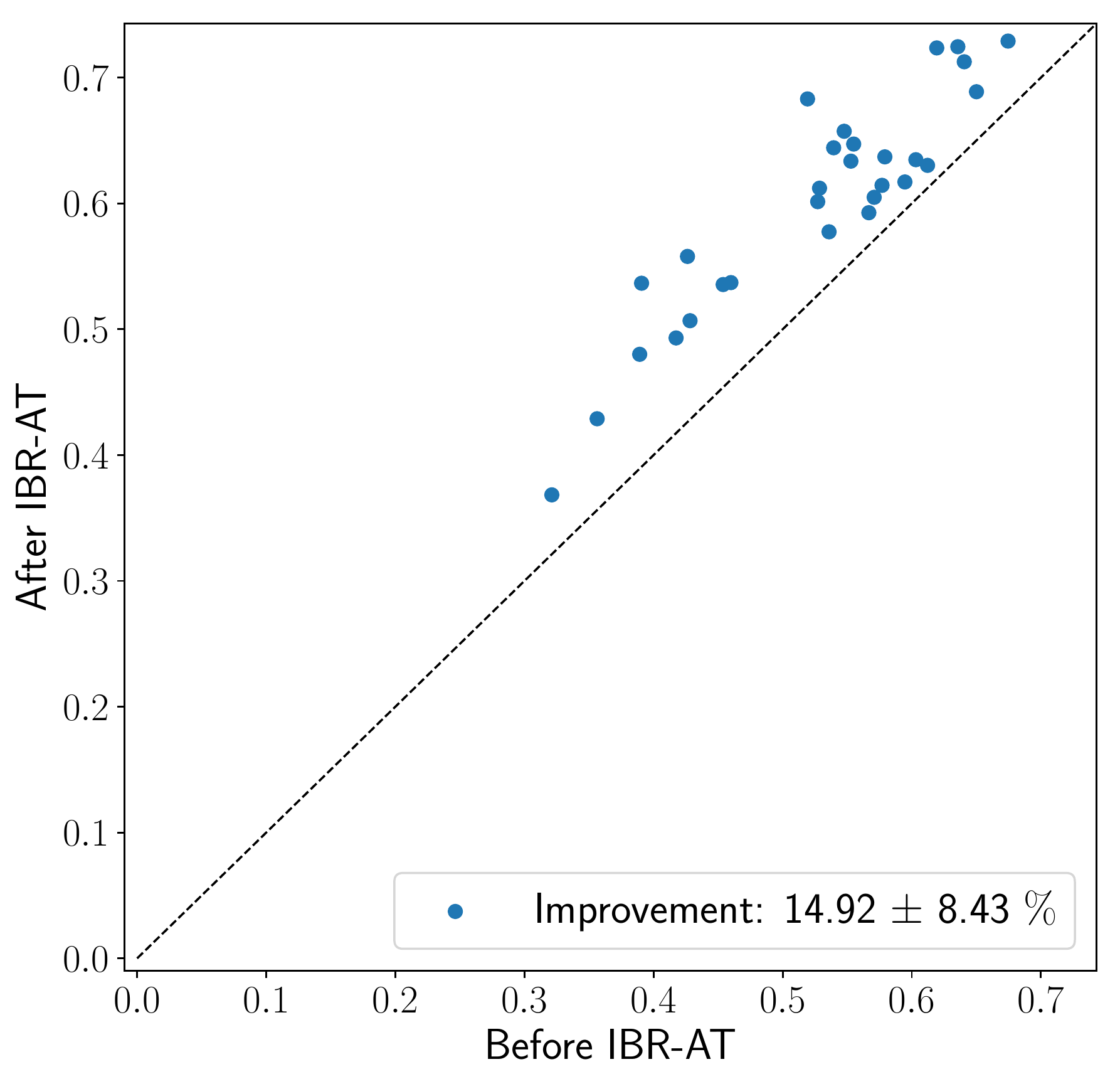}}
    \subfigure[]{\includegraphics[width=0.32\textwidth]{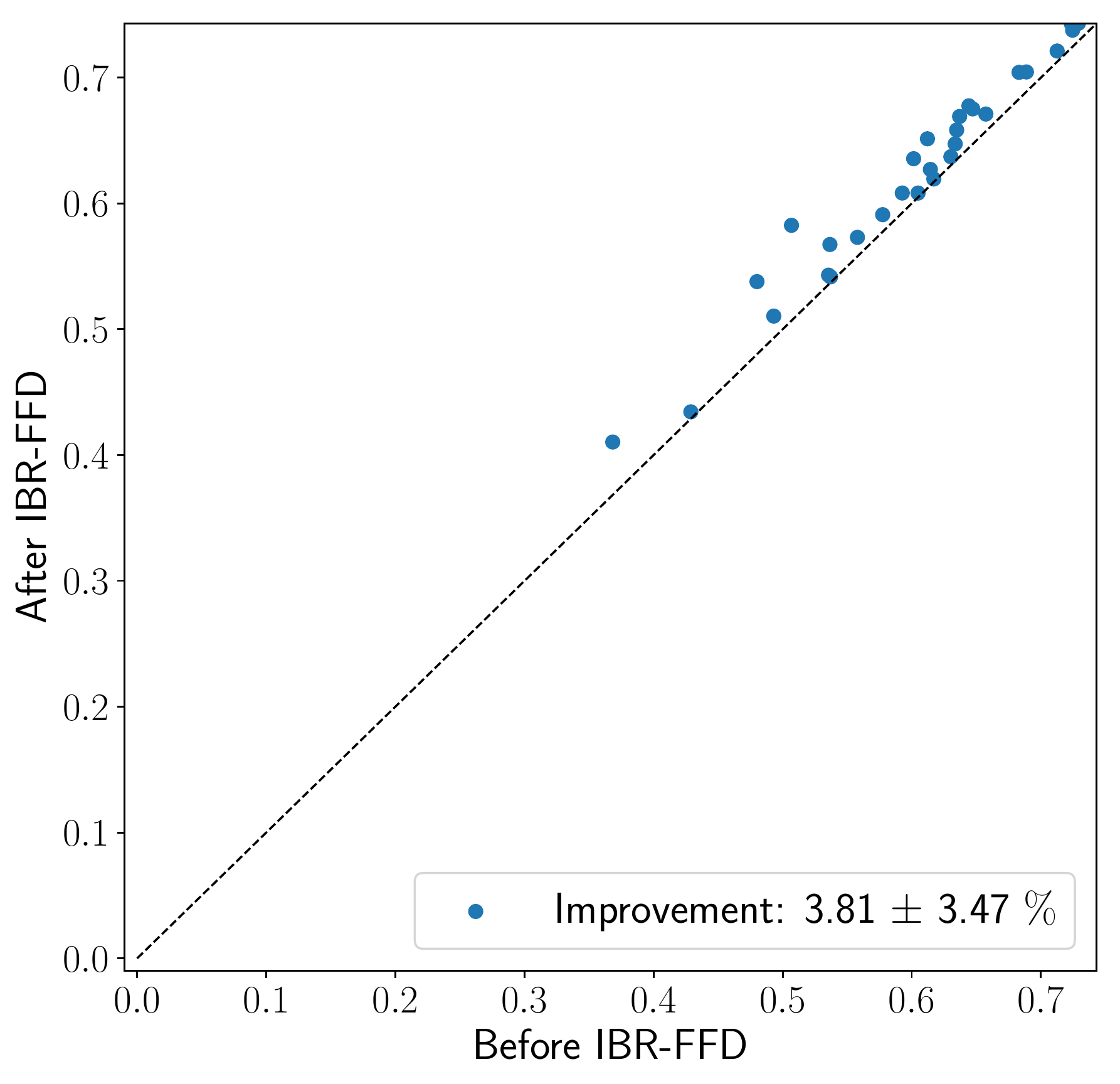}}
    \subfigure[]{\includegraphics[width=0.32\textwidth]{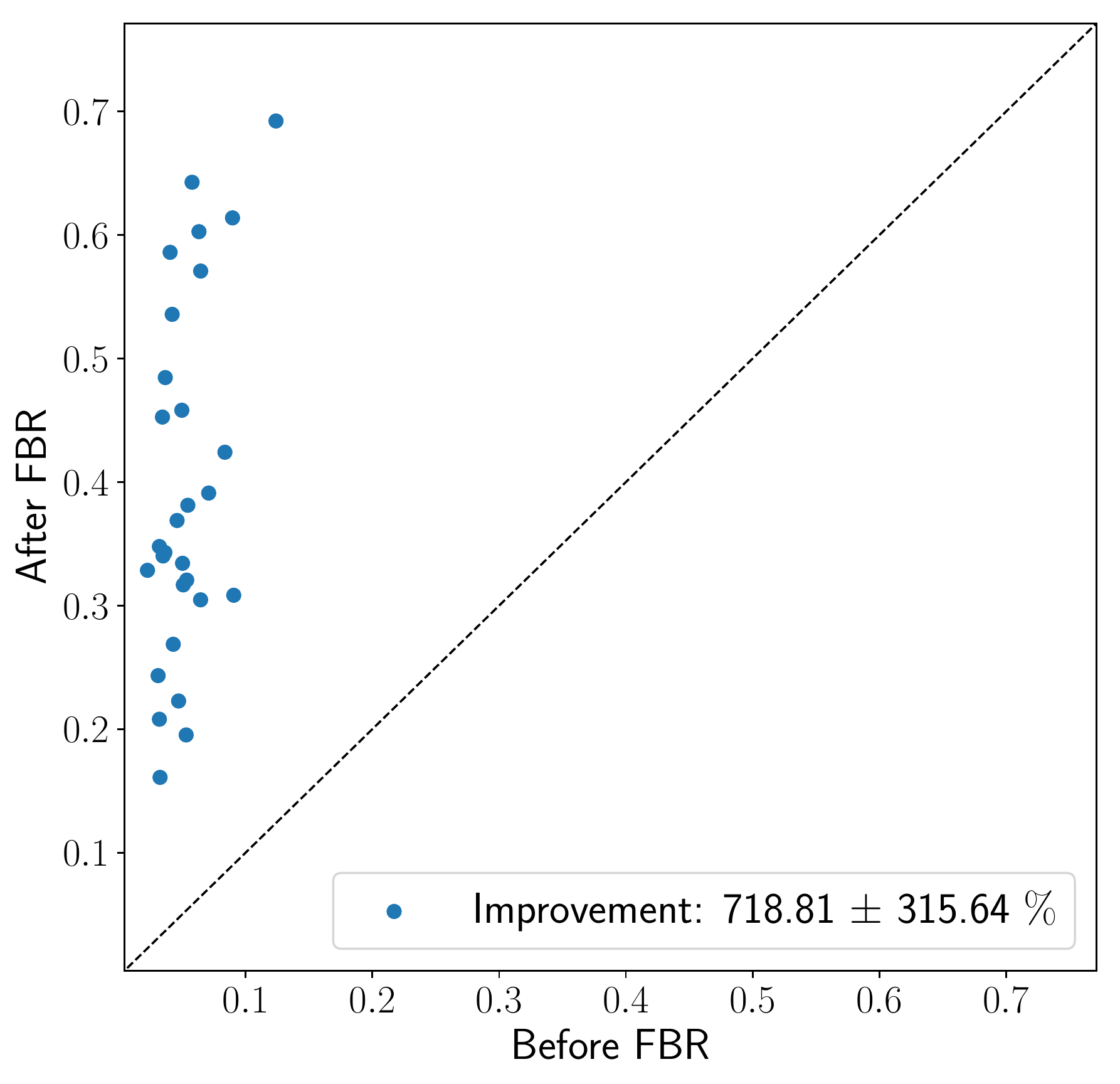}}
    \subfigure[]{\includegraphics[width=0.32\textwidth]{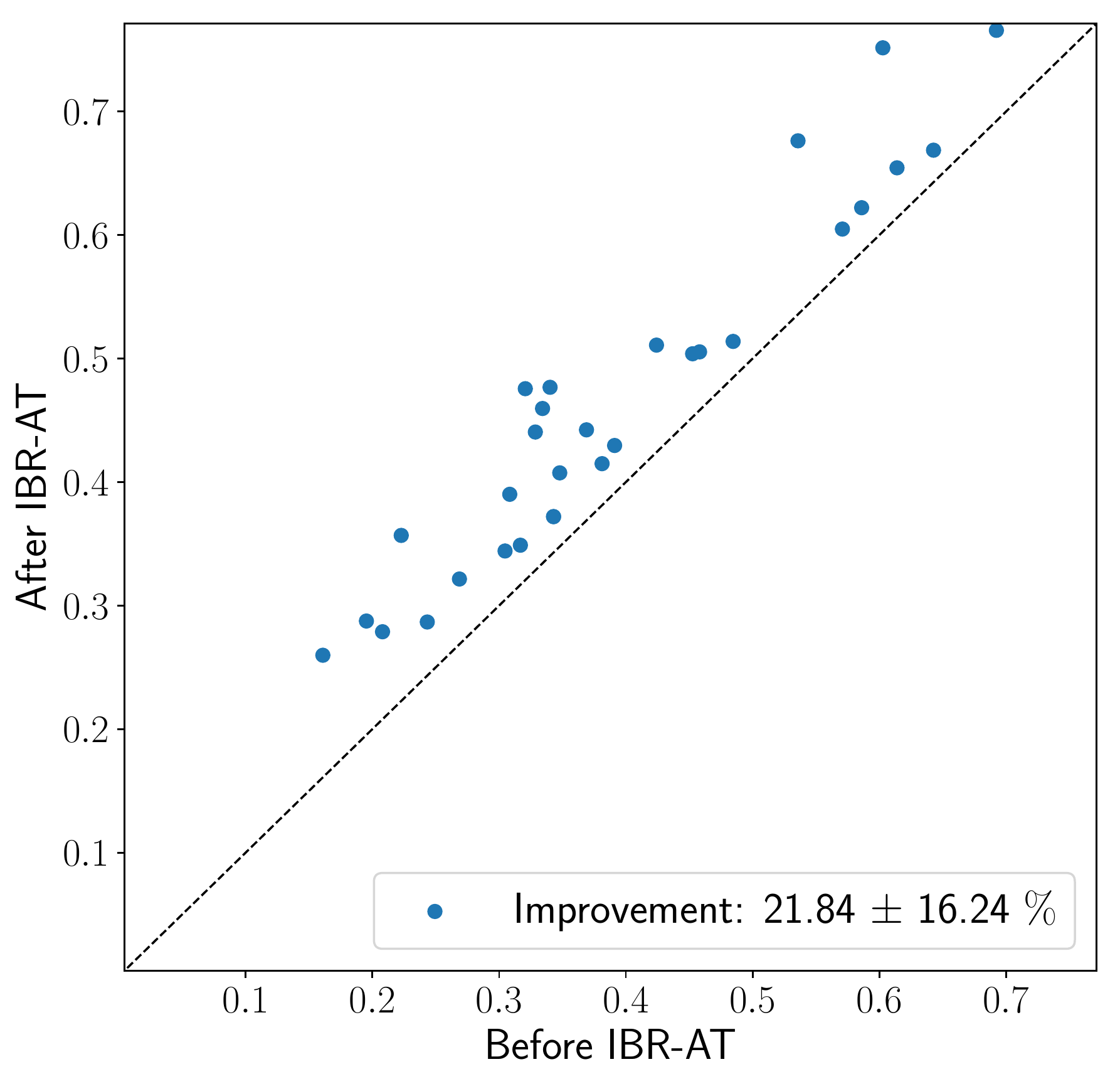}}
    \subfigure[]{\includegraphics[width=0.32\textwidth]{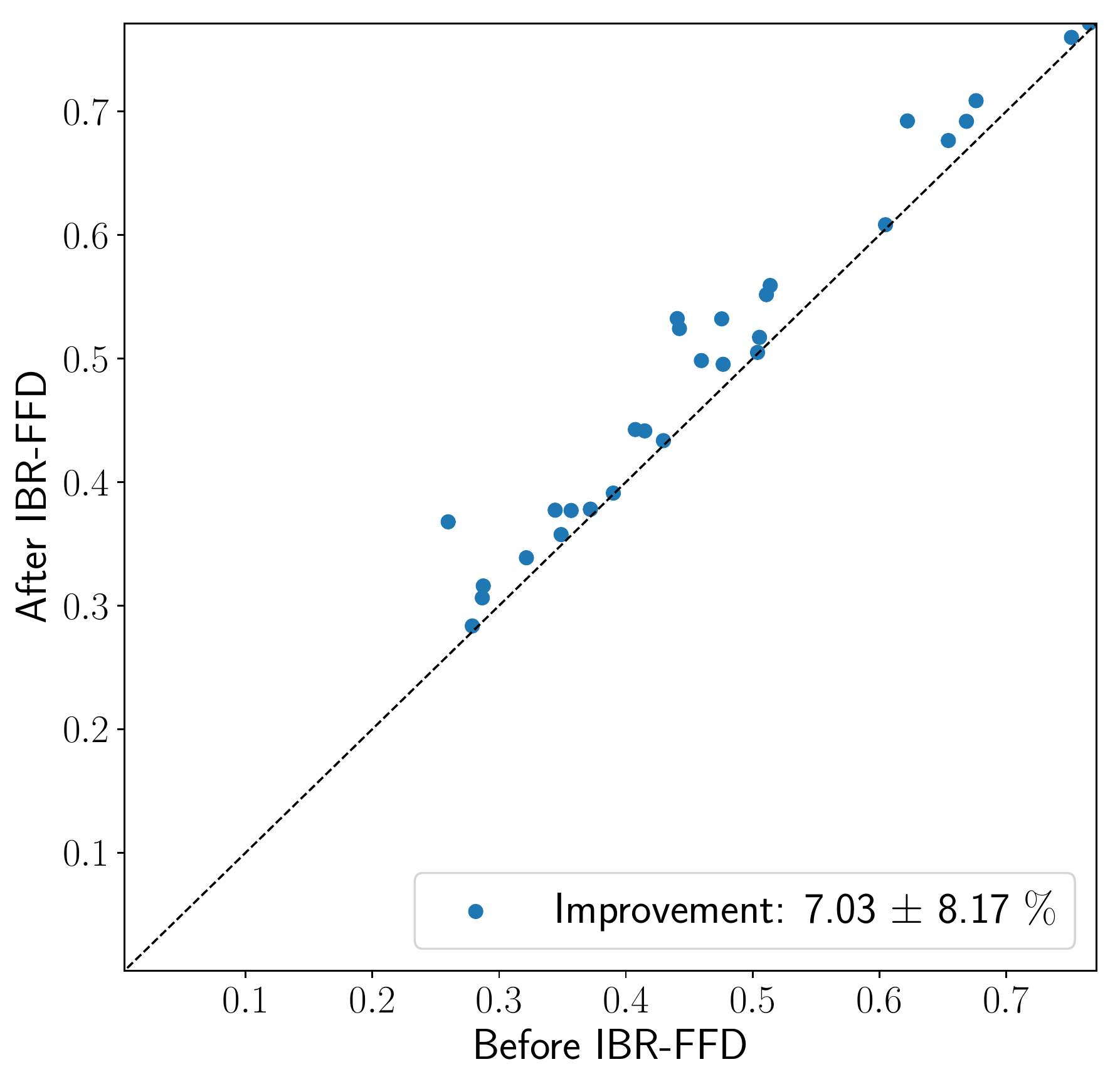}}
    \caption{Scatter plots of the VE-NCC values before and after each step of the proposed hybrid method and percentage of improvement for each step: (a)(b)(c) healthy cases: (d)(e)(f) pathological cases.}
    \label{improvement}
\end{figure}

Additionally, we performed a more in-depth analysis of the impact of the different steps in the proposed hybrid configuration. Figure \ref{improvement} depicts scatter plots of the VE-NCC values before and after each step of the hybrid method for both healthy and pathological cases. Also, the percentage of improvement obtained with each step is computed and shown in the plots. It is observed that the biggest contribution comes from the initial FBR. The improvement decreases with each step as minor adjustments in the estimated transformation are required. The presence of pathologies in the images does not affect the overall behavior of the proposed hybrid method, as similar conclusions can be drawn from sets of both scatter plots. However, the average VE-NCC values are slightly lower for the pathological cases, at the same time that the variance is slightly higher. This is an indication of the slight influence of the pathological structures in the VE-NCC. The maximum value is not necessarily the same for every image pair, although this does not affect the optimal transformation and a successful registration or the retinography/angiography pairs is achieved for all the cases, regardless of being pathological or not.

\section{Conclusions} \label{conclusions}
The joint analysis of color fundus retinography and fluorescein angiography usually requires the registration of the images. In this work, a hybrid method for the multimodal registration of pairs of retinographies and angiographies is presented. Domain-specific solutions, exploiting the presence of the retinal vasculature in both image modalities, were proposed for both the feature-based and intensity-based registration steps of the hybrid method proposed in this work. The use of a domain-adapted similarity metric allows the estimation of high order transformations that increase the accuracy of the registration. Simultaneously, an accurate registration is only feasible departing from the initial registration with domain-specific landmarks. Different experiments were conducted to validate the suitability of the proposed method and to evaluate the contribution of each registration step. The results demonstrated that the hybrid method outperforms the individual application of each of its constituent approaches, and each composition subset of them, highlighting the suitability of the proposal in the multimodal registration.

\section*{Acknowledgments}
This work is supported by Instituto de Salud Carlos III, Government of Spain, and the European Regional Development Fund (ERDF) of the European Union (EU) through the \mbox{DTS15/00153} research project, and by the Ministerio de Economía, Industria y Competitividad, Government of Spain, through the $~$ \mbox{DPI2015-69948-R} research project. The authors of this work also receive financial support from the European Regional Development Fund (ERDF) and European Social Fund (ESF) of the EU, and the Xunta de Galicia through Centro Singular de Investigación de Galicia, accreditation 2016-2019, ref. ED431G/01 and Grupo de Referencia Competitiva, ref. ED431C 2016-047 research projects, and the predoctoral grant contract ref. ED481A-2017/328.

\bibliographystyle{splncs}
\bibliography{references.bbl}

\end{document}